# Adaptive Blind Watermarking Using Psychovisual Image Features

Arezoo PariZanganeh[1], Ghazaleh Ghorbanzadeh[1], Zahra Nabizadeh ShahreBabak[1],
Nader Karimi[1], Shadrokh Samavi[1,2,3]

[1]Department of Electrical and Computer Engineering, Isfahan University of Technology, 84156-83111, Iran
[2]Department of Electrical and Computer Engineering, McMaster University, L8S 4L8, Canada
[3]Department of Computer Science, Seattle University, Seattle, 98122 USA

*Abstract*— With the growth of editing and sharing images through the internet, the importance of protecting the images' authority has increased. Robust watermarking is a known approach to maintaining copyright protection. Robustness and imperceptibility are two factors that are tried to be maximized through watermarking. Usually, there is a trade-off between these two parameters. Increasing the robustness would lessen the imperceptibility of the watermarking. This paper proposes an adaptive method that determines the strength of the watermark embedding in different parts of the cover image regarding its texture and brightness. Adaptive embedding increases the robustness while preserving the quality of the watermarked image. Experimental results also show that the proposed method can effectively reconstruct the embedded payload in different kinds of common watermarking attacks. Our proposed method has shown good performance compared to a recent technique.

## I. INTRODUCTION

With the increasing ease of access to digital multimedia and the rapid growth of using media editing and sharing software, the problem of copyright protection has been more challenging. In the literature on information security, digital watermarking technology is a vast field of interest for scholars to avoid unauthorized copying and manipulation of digital content [1, 2]. Robust watermarking is one of the popular watermarking fields trying to preserve the watermark content during different attacks that are possible in real-world conditions [3]. Some of these attacks on a watermarked image are additive Gaussian noise, median, and Gaussian filtering, lossy image compression, and geometric attacks such as rotation and cropping [1].

Imperceptibility and embedding capacity are two other criteria that should be considered simultaneously. Imperceptibility indicates the similarity of the cover image before and after embedding the watermark. Embedding capacity defines the number of bits the proposed algorithm can embed in the cover image. These three criteria make a triangle trade-off that is needed to be managed based on conditions of use [3, 4]. Embedding capacity defines the number of bits the proposed algorithm can embed in the cover image. Furthermore, based on the limitations of hardware and computation resources in conditions of use, the complexity of the embedding algorithm can be another factor to be dealt with [2].

Watermarking techniques can be compared and classified based on different aspects. According to the domain that embedding is applied in, watermarking techniques can be divided into spatial and transform domain watermarking algorithms. Spatial watermarking algorithms directly change pixel values to embed the watermark. For instance, in the early watermarking methods, the least significant bits of pixel values were substituted based on the stream of watermark bits [3].

Although spatial methods provide simple implementation and need lightweight computations, the transform domain-based algorithms can offer a more robust watermarking scheme. Discrete Fourier Transform (DFT), Discrete Cosine Transform (DCT), and Wavelet Transform (WT) are some of the most commonly used transforms in this field [5]. DCT is used in many works based on the wonderful compression of energy of the signal in low frequencies that it performs. Franwan et al. [6] modified middle-frequency DCT coefficients based on a psychovisual threshold. Wavelet transform is widely used because of proper time and frequency decomposition characteristics. It is also used in JPEG2000, which makes it more resistant to compression attacks. Zue et al. [7] proposed a watermarking algorithm based on Integer WT (IWT) combined with Singular Value Decomposition (SVD). SVD is applied to the low-frequency coefficients of IWT, and then the first singular value is quantized. The quantization step is optimized with the help of a genetic algorithm. In this way, the robustness and imperceptibility of the watermarking algorithm are increased. Naseem et al. [8] also used IWT as the transform domain. In this method, the selection of blocks to embed payload bits are determined based on a Fuzzy Rule-Based System (FRBS). The coefficients in which the payload is hidden are likewise determined based on the proposed FRBS.

Other types of transforms are presented to compensate for the drawbacks of the previously mentioned transform domains. For example, Hu et al. [4] addressed the deficiency of previous methods in the geometric attacks by using Contourlet Transform (CT). Based on the more efficient representation of two-dimensional discontinuities of CT in comparison with WT and its variants, CT is chosen as the transform domain in this work. In another work, Hu et al. [9] presented a watermarking algorithm that embedded payload into low-order Zernike moments and showed that their method is more robust against rotation and scaling attacks.

There are a large number of researches combining different transform domains and optimization methods to yield high robustness while keeping the fidelity of the watermarked images the most that possible. For example, Devi et al. [10] applied an extension of WT while the scaling factor in their method was optimized based on an optimization method called the hybrid grasshopper–BAT algorithm. The techniques in [11-13] applied combinations of a variation of WT and DCT as the embedding domain and increased the robustness using the SVD algorithm. Another way to increase the robustness of watermarking is to embed the watermark redundantly. In applications where the capacity is of lower priority, redundancy can effectively make the watermarking algorithms resistant to different image processing manipulations. In [14], block-wise redundancy is applied in embedding, and voting is utilized in extraction in a DCT-based watermarking algorithm. In [5, 15], redundancy is used to improve robustness during an adaptive strength factor for each block to increase the fidelity of watermarked image to the original host image. In [15] saliency map is used as an auxiliary input, and a fuzzy system has been presented, which increased the computation complexity of the method. The method in [5] also needs global information on image blocks to determine the embedding threshold.

In this work, we have introduced an approach that achieved better results than the method in [14] regarding robustness and imperceptibility. Motivated by the benefits of the adaptive embedding method presented in [15], we presented an adaptive strength factor calculation method. For this purpose, edge information of the cover image and its brightness are utilized to determine the embedding strength in each cover image block. A combination of DCT and two-level DWT is utilized as the watermarking transform domain. Block-based redundancy is also applied to improve robustness. The results of the proposed method are compared with the method in [14] in terms of the Peak Signal-to-Noise Ratio (PSNR) and Structural Similarity Index (SSIM) of watermarked images compared with the original ones. Bit Error Rate (BER) and Normalized Cross-Correlation (NC) values of the reconstructed watermark are also presented. The evaluation metrics are reported for four standard classic watermarking images. In calculating each criterion, 20 randomly created watermarks are embedded in the host image, and the averaged value is mentioned for a more comprehensive comparison.

The remainder of this paper is organized as follows. Section II describes our proposed method in its different parts. Experimental results are presented in section III. Finally, section IV concludes the paper

## II. PROPOSED METHOD

In this section, first, we explain the method of calculating the Strength Factor (SF) for each block and then investigate the basic concept of the DCT transformation, which we know as a widely used transformation in the watermark. In the end, embedding and extracting algorithms are explained.

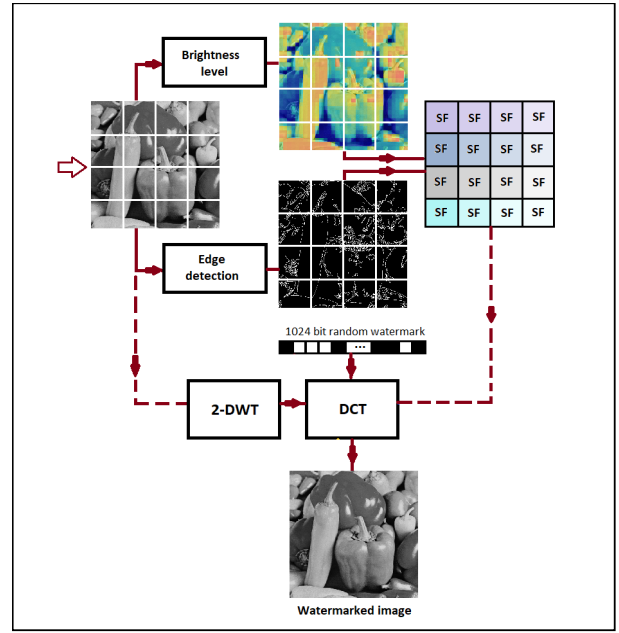

Fig. 1. The proposed embeding procedure.

### A. Strength Factor Determination

Our studies show that images with more edges are more suitable for data embedding. Making changes in this part is less recognized by human vision. Therefore, we used the canny edge detection filter to extract the edges of the cover image. We calculate the Strength Factor (SF) based on the number of edges obtained in each image block. We assign a small SF to blocks with fewer edges and a large SF to those with more edges. The image brightness level is another factor used to determine the Strength Factor in our method. According to our experiments, using more intense SF in parts of the image with a lower brightness level effectively maintains the image's visual quality. We use the normalized values of the edge and the brightness level of the image blocks to calculate the final SF. Coefficients have been determined by considering the quality of the watermarked image and its resistance to different attacks.

### B. Basic Concept

It is common to use a combination of DWT and DCT transformations in watermarking methods. The selection of the number of DWT transform levels and frequency sub-levels for watermarking is varied in different methods.

Basically, in the methods that use DCT transformation, the entire image or smaller blocks are transferred by basic conversion coefficients from the spatial domain to the frequency domain. As mentioned in [16] and previous studies in this field, the obtained coefficients are divided into high, low, and middle-frequency bands. Different studies show that using mid-frequency coefficients is more appropriate for robustness and transparency. Of course, determining the proper threshold, choosing the best coefficients, and determining the mechanism of moving the coefficients in various articles are different, which have also provided different results. But the common point in most of the presented methods is that the value of the coefficients is used for comparison.

**Algorithm 1.** Proposed DCT watermark embedding procedure.

**Input:** Coefficients $C[v,u]$ and $C[u,v]$, Watermark
**Output:** coefficients $C'[v,u]$ and $C'[u,v]$
**Begin**
   if Watermark == 0:
    if $C[v,u] - C[u,v] > T$ & both are same sign:
      without change
    else if $C[u,v] - C[v,u] > T$ & both are same sign:
      $C[v,u] = abs(C[v,u])/2$
      $C[u,v] = -1 * abs(C[u,v])/2$
    else:
      $C[v,u] = abs(C[v,u])$
      $C[u,v] = -1 * abs(C[u,v])$
    end if

   if Watermark == 1:
    if $C[u,v] - C[v,u] > T$ & both are Same sign:
      without change
    else if $C[v,u] - C[u,v] > T$ & both are Same sign:
      $C[u,v] = abs(C[u,v])/2$
      $C[v,u] = -1 * abs(C[v,u])/2$
    else:
      $C[u,v] = abs(C[u,v])$
      $C[v,u] = -1 * abs(C[v,u])$
    end if
   end if
**End**

*C. Embedding Process*

The embedding steps in the presented method are illustrated in Fig. 1 and detailed as follows:
Firstly, we extract the brightness level map and the image edge map by applying the canny filter on the cover image. After that, the number of pixels in each block that are determined as an edge by a Canny filter, and the sum of the brightness in each block is calculated. In (1), the calculation of the Strength Factor from edge and brightness is shown. Edge_count is the number of edges in a block, and the brightness_level defines the block brightness. In the next step, we Apply two-level DWT on 128×128 non-overlapping blocks of the image. LH1 and HL1 sections with the size of 64×64 from the first level and LH2, HL2, and HH2 sections with the size of 32×32 from the second level are selected to make 8×8 blocks. Then we apply DCT on each block to choose appropriate coefficients $C[v,u]$ and $C[u,v]$. We calculate SF based on the edge map and brightness level map for each 128×128 block.

$$SF = |\alpha \times Edge\_count - \beta \times Brightness\_level| \quad (1)$$

We use (2) to calculate the comparison threshold using.

$$T = SF \times (|C[v,u]| + |C[u,v]|) + 0.001 \quad (2)$$

We apply the necessary changes according to Algorithm 1 to produce coefficients $C'[v,u]$ and $C'[u,v]$, based on the value of the watermark, the difference of the coefficients and their sign. In the end, we apply a DCT inverse for each 8×8 block, then use a two-level DWT inverse for 128×128 blocks to make the watermarked image.

*D. Extracting Process*

Considering that the presented method is blind, the watermark can be extracted by receiving the watermarked image without additional information. Data extraction steps are as follows. Firstly, the cover image is divided into blocks of 128×128. Then, two levels of DWT are applied on each block separately. LH1 and HL1 sections from the first level and LH2, HL2, and HH2 sections from the second level are selected as the output of the DWT block. Each of these maps is then fragmented into 8×8 blocks, and DCT is applied on each block. Finally, we used Algorithm 2 to extract the embedded bit in blocks.

**Algorithm 2.** Extracting data process.

**Input:** Coefficients $C'[v,u]$ and $C'[u,v]$
**Output:** Watermark
**Begin**
   if $C'[v,u] > C'[u,v]$ & both are same sign:
    $W = 0$
   else if $C'[u,v] > C'[v,u]$ & both are same sign:
    $W = 1$
   else if $C'[v,u]$ is negative:
    $W = 1$
   else $C'[u,v]$ is negative:
    $W = 0$
   end if
**End**

III. EXPERIMENTAL RESULTS AND EVALUATION

We evaluated the proposed method's effectiveness with several tests in this study. We use grayscale images of size 512×512 as the cover images, and a randomly generated 256-bit string is used as the watermark. We use PSNR and SSIM to evaluate the imperceptibility of the watermark and NC and BER to check the robustness of the watermark against different attacks.

In this method, it is also possible to add watermark data with redundancy. Experiments show that using coefficients LH1 and HL1 sections from the first level of DWT conversion and coefficients LH2, HL2, and HH2 sections from the second level of conversion can be a good choice. For example, in one test, we import the watermark data eleven times into the image and at the specified coefficients. Then, by voting on the obtained results, we get the final watermark data. Also, coefficients [6,4] and [4,6] are used in DCT conversion. These coefficients which are on the border of high-frequency and mid-frequency coefficients, keep the watermark image resistant to attacks.

TABLE 2. IMPERCEPTIBILITY OF OUR PROPOSED METHOD IN BOTH NON-ADAPTIVE AND ADAPTIVE SCHEMES VS. THE METHOD IN [14]

| Method | | Test Images | | | |
|---|---|---|---|---|---|
| | | *Lena* | *Baboon* | *Peppers* | *Goldhill* |
| Non-adaptive scheme | PSNR | **46.02** | **42.80** | **46.98** | **45.88** |
| | SSIM | 0.99 | 0.99 | 0.99 | 1 |
| Our proposed method | PSNR | **46.57** | **43.23** | **47.11** | **46.75** |
| | SSIM | 1 | 0.98 | 1 | 0.99 |
| Method in [14] | PSNR | 45.19 | - | 46.34 | 46.00 |
| | SSIM | 1 | - | 1 | **1** |

TABLE 1. NC RESULTS OF OUR PROPOSED METHOD IN DIFFERENT TYPES OF ATTACKS

| Attack Type | | Test Images | | | |
|---|---|---|---|---|---|
| | | *Lena* | *Baboon* | *Peppers* | *Goldhill* |
| Without Attacks | | 1 | 1 | 1 | 1 |
| MF | Non-adaptive | 0.80 | 0.86 | 0.69 | **0.85** |
| | Adaptive | **0.84** | **0.92** | **0.78** | 0.83 |
| | Method in [14] | 0.68 | 0.81 | 0.67 | 0.82 |
| S&P | Non-adaptive | 0.77 | 0.97 | 0.81 | 0.88 |
| | Adaptive | **0.78** | **0.98** | **0.82** | **0.89** |
| | Method in [14] | 0.75 | 0.97 | 0.80 | **0.89** |
| HE | Non-adaptive | 0.97 | 0.99 | **0.97** | 0.95 |
| | Adaptive | **0.99** | **1** | 0.96 | **0.96** |
| | Method in [14] | 0.97 | **1** | 0.96 | **0.96** |

We have used four images of Lena, Baboon, Peppers, and Goldhill for comparison with [14]. The calculated PSNR values are reported in Table 1. As mentioned, we have used an edge map and image brightness level to determine SF. In this study, instead of considering the sign of DCT coefficients for moving them, their signs and values are used. To see the effect of this change, we use a fixed Strength Factor to calculate the threshold. The results are shown as a Non-Adaptive method. Both proposed non-adaptive and adaptive methods have better PSNR values, while the method in [14] has slightly outperformed our strategies in terms of SSIM.

Various attacks have been applied to the image, and their NC values have been calculated. The experiments evaluate the robustness of the proposed watermarking method. Table 2 reports some of these results, such as median filter (MF), salt and pepper noise (S&P), and histogram equalization (HE). It can be seen that, in most of the results, using the adaptive approach has increased the NC of the reconstructed watermark compared with the original one.

Furthermore, Fig.2 and 3 are graphs comparing Gaussian attacks and JPEG compression. The blue bars in Fig.2 determine the NC values of our proposed adaptive method. It can be seen that our proposed method has outperformed the non-adaptive method and also the method in [14]. For Gaussian noise with variances of 0.003 and 0.005, our proposed non-adaptive method also performs better than the technique of [14]. Due to the lack of BER results from Goldhill in [14], we only report the BER of our proposed method with the other three images.

## IV. CONCLUSION

In this paper, an adaptive robust watermarking algorithm is proposed that operates based on a combination of two-level wavelet transform and DCT. The applied DCT is different from the basic version, and the method of moving the coefficients is changed. The proposed method takes advantage of edge and brightness information in each block of the host image to determine the strength factor of embedding. It also redundantly embeds the watermark bit stream into the host image to promote the resistance of the algorithm in different image processing attacks. The experimental results demonstrate that the proposed method is imperceptible compared with the comparable watermarking algorithms and, at the same time, is robust against a wide range of common watermarking attacks.

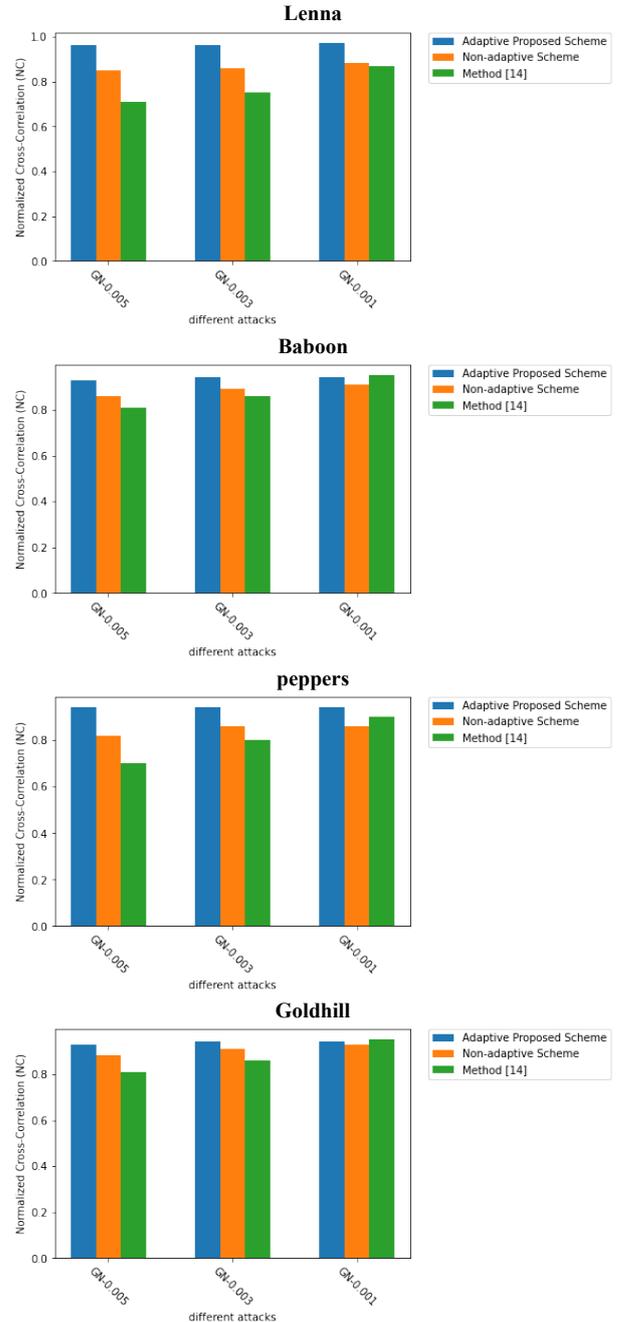

Fig. 2. Normalized Cross-Correlation (NC) results of Gaussian Noise (GN) with different variances for our proposed method in both non-adaptive and adaptive schemes vs. the method in [14]

Furthermore, because of its low computational complexity, it also can be applied to frame sequences of videos.

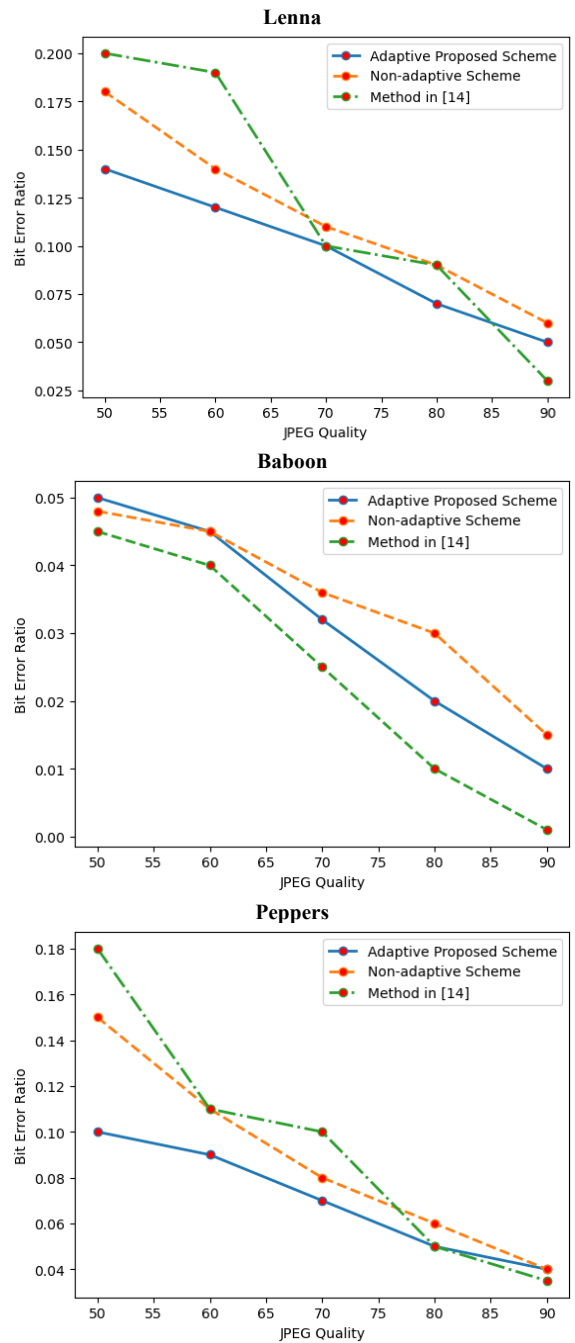

Fig. 3. Bit Error Rate (BER) results of JPEG copmression attack for our proposed method in both non-adaptive and adaptive scheme vs. the method in [14].